\pdfoutput=1
\documentclass[11pt,a4paper]{article}
\usepackage[hyperref]{emnlp2020}
\usepackage{times}
\usepackage{latexsym}

\usepackage{microtype}

\usepackage{multirow}
\usepackage{amsmath}
\usepackage{amsthm}
\usepackage{amssymb}
\usepackage{graphicx}

\aclfinalcopy 

\title{Beyond [CLS] through Ranking by Generation}

\author{Cicero Nogueira dos Santos, Xiaofei Ma, \\
\textbf{Ramesh Nallapati, Zhiheng Huang and Bing Xiang} \\
  AWS AI \\
  New York City, NY \\
  {\tt \{cicnog,xiaofeim,rnallapa,zhiheng,bxiang\}@amazon.com}}

\date{}

\begin{document}
\maketitle
\begin{abstract}
Generative models for Information Retrieval, where ranking of documents is viewed as the task of generating a query from a document's language model, were very successful in various IR tasks in the past. However, with the advent of modern deep neural networks, attention has shifted to discriminative ranking functions that model the semantic similarity of documents and queries instead. Recently, deep generative models such as GPT2 and BART have been shown to be excellent text generators, but their effectiveness as rankers have not been demonstrated yet. 
In this work, we revisit the generative framework for information retrieval and show that
our generative approaches are as effective as state-of-the-art semantic similarity-based discriminative models for the answer selection task.
Additionally, we demonstrate the effectiveness of unlikelihood losses for IR. 
\end{abstract}

\section{Introduction}
Most recent approaches for ranking tasks in Information Retrieval (IR) such as passage ranking and retrieval of semantically related questions have focused primarily on discriminative methods using neural networks that learn a similarity function to compare questions and candidate answers \cite{severyn:2015,dossantos2015,tan2016,tay2017:yqa,tay2018:wsdm}.
On the other hand,
classical literature on probabilistic models for IR showed that language modeling, a type of simple generative model, can be effective for document ranking \cite{zhai:2008,zhai_lafferty,ponte_croft}.
The key idea consists of first training a unique language model $lm_i$ for each candidate document $d_i$, then using the likelihood  of generating the input query using $lm_i$, denoted by $P(q|lm_i)$, as the ranking score for document $d_i$.

Recent advances in neural language models (NLMs) have led to impressive improvements in the quality of automatically generated text \cite{radford2019language}.
However, to the best of our knowledge, there is no existing work
 on exploring the effectiveness of modern generative models such as GPT2, for complex ranking tasks such as answer selection.
In this work, 
we intend to fill this gap by demonstrating that large pretrained generative models can be very effective rankers.
Unlike classic LM based approaches for IR that employ separate LMs for each document,
our proposed method uses a single global LM that applies to all documents.
The global pretrained generator is fine-tuned on the task of query generation conditioned on document content as the context.
Additionally, 
in order to leverage both positive and negative examples,
we propose the use of (1) unlikelihood loss on negative examples and (2) ranking loss on the likelihood of positive and negative examples.
At inference time, given an input query,
our method scores each candidate document using the likelihood of generating the query given the document, as estimated by our fine-tuned global LM.

\begin{figure*}[t]
\centering
    \includegraphics[width=1\textwidth]{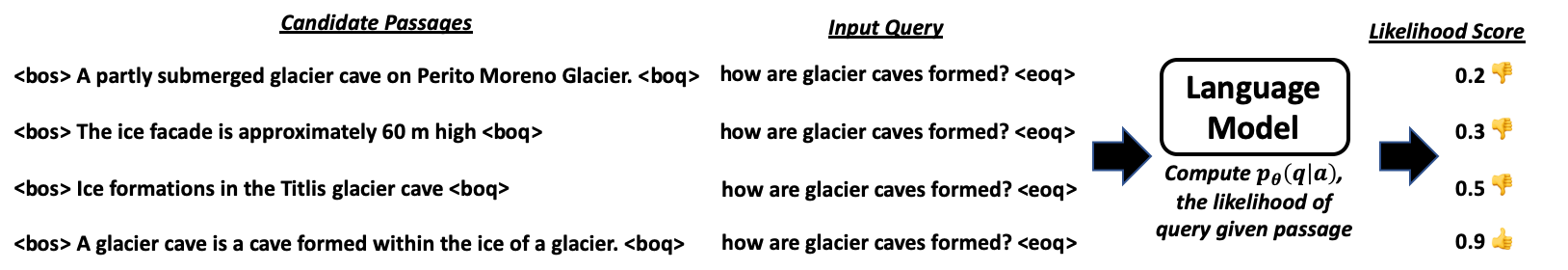}
    \caption{{\small Illustration of the inference step of our \emph{ranking by generation} approach. Each candidate passage $a_k$ is ranked based on the likelihood of generating the question $q$ conditioned on the passage, $p_{\theta}(q|a_k)$.}}
    \label{fig:lmranker}
\end{figure*}

We focus our experiments on the task of answer selection (a.k.a passage ranking).
In this task, given an input question and a set of candidate passages,
the goal is to rank the candidate passages so that passages containing the correct answer appear at the top of the ranked list.
Considerable body of work exists on the use of NNs for this task \cite{feng2016:insqa,severyn:2015,tan2016,santos2016attentive,jinfeng2016,wang17:bilateral},
where the most recent ones use BERT-based models that perform discrimination based on the special $[CLS]$ token \cite{nogueira2019passage,li2019:bertsel,xu2019:weaksup}.
A contemporaneous work by \citet{nogueira2020document} also proposes a generative approach for the passage ranking task. However, while their approach decides the relevance of a passage by generating a single keyword (e.g. \emph{true} or \emph{false}), our method uses the conditional likelihood of generating the question given the passage as a relevance score.

We perform extensive experiments using GPT2 \cite{radford2019language} and BART \cite{lewis2019bart},
which are Transformer-based LMs \cite{vaswani:NIPS2017} that were pretrained using large volumes of textual data.
The LMs are fine-tuned on four different passage ranking datasets separately:
WikipassageQA,
WikiQA,
InsuranceQA\_V2,
and YahooQA.
Our experimental results indicate that our generative approaches are as effective as state-of-the-art discriminative-based approaches for answer selection.

\section{Ranking by Generation}
\subsection{Background}
The goal in language modeling is to learn the probability distribution $p(x)$ over variable-length token sequences $x=(x_1,x_2,...,x_{|x|})$,
where the tokens come from a fixed size vocabulary, $x_i \in V$.
When training an LM with {\it causal language modeling objective}, which consists of predicting the next token by looking at the past only, 
we can denote this distribution by the conditional probability of the next token given the previous ones \cite{bengio:2003}:
\begin{equation}
    p(x)= \prod^{|x|}_{i=1}{p(x_i|x_{<i})}
\end{equation}

GPT2 \cite{radford2019language} is an example of a state-of-the-art neural LM trained with causal language modeling objective. The usual approach to train an LM using a neural network with parameters $\theta$
consists on performing maximum likelihood estimation (MLE) by minimizing the negative log-likelihood over a large text corpus $D=\{x^1, x^2, ... , x^{|D|}\}$, where each $x_k$ is a document of length $|x_k|$: 
\begin{equation}
    \mathcal{L}(D)= -\sum^{|D|}_{k=1} \sum^{|x^k|}_{i=1}{\log p_{\theta}(x_i^k|x_{<i}^k)}
\label{eq:MLE}
\end{equation}

Conditional LMs are a simple extension of regular LMs where the generation is conditioned on some additional context $c$ \cite{keskar2019ctrl}:
\begin{equation}\label{eq:conditional_lm}
    p(x|c)= \prod^{|x|}_{i=1}{p(x_i|x_{<i},c)}
\end{equation}

\subsection{Proposed Ranking Approach}
Our proposed approach for passage ranking by generation consists of first fine-tuning a pretrained large LM on the task of question generation conditioned on the passage, using the conditional LM approach shown in Eq. \ref{eq:conditional_lm}.
In practice, 
each input for the fine-tuning step is as follows:
\begin{center}
\begin{footnotesize}
\begin{verbatim}
<bos> passage <boq> question <eoq>
\end{verbatim}
\end{footnotesize}
\end{center}

\noindent where the \emph{passage} is considered as a prompt,
and the log-likelihood used in the training comes only from the tokens starting after the keyword \emph{$<$boq$>$}, since we use the passage as a conditioning context. In other words, at training time, we minimize the negative conditional log-likelihood $-\log P(q|a)$, where $a$ is a passage relevant to the query $q$. At inference time,
given a query $q$, our conditional LM scores each candidate passage $a_k$ using the likelihood of generating the question conditioned on the passage, $s(a_k)=p_{\theta}(q|a_k)$.
Fig. \ref{fig:lmranker} illustrates the inference step of our proposed approach.

\subsection{Unlikelihood Loss for Ranking}
Datasets for training passage rankers normally contain both positive and negative examples.
Therefore, it is natural to use both types of examples in order to leverage all the available data.
Let $D$ be the set of examples $(q,a,y)$, where $y$ is $1$ if the passage $a$ is a positive answer for $q$, or $0$ otherwise. 
We fine-tune the LM using the following loss function:
\begin{equation}
\begin{aligned}
    \mathcal{L}(D)= & -\sum_{(q,a,y) \in D}{\sum^{|q|}_{i=1}{y\log(p_{\theta}(q_i|q_{<i},a))}} \\
    & + (1-y)\log (1-p_{\theta}(q_i|q_{<i},a))
\label{eq:LUL}
\end{aligned}
\end{equation}

The second term in Eq. \ref{eq:LUL} resembles the unlikelihood training objective of \citet{welleck2019neural}.
However, while we use an unlikelihood objective with the aim of teaching the LM which questions are unlikely given the passage,
\citet{welleck2019neural} use an unlikelihood objective with the aim of improving text generation.
We use the acronym $LUL$ to refer to the loss function in Eq. \ref{eq:LUL}, which performs likelihood and unlikelihood estimation.

We experimented an additional loss function to fine-tuning the LMs 
which consists on imposing a pairwise ranking loss on the likelihood ($RLL$) of positive and negative examples as follows:
\begin{eqnarray}
    \mathcal{L}(D) &= & \sum_{(q,a+,a-) \in D} \max\{0, \lambda - \log p_{\theta}(q|a^+) \nonumber\\
    & + & \log p_{\theta}(q|a^-)\} 
\label{eq:RLL}
\end{eqnarray}

The use of unlikelihood losses to penalize negative examples is a natural choice for fine-tuning generative models. Note that Eq. \ref{eq:LUL} is an extension of the regular cross-entropy loss where we just added the unlikelihood term, while Eq. \ref{eq:RLL} is its ranking-based (hinge loss) version. 
The unlikelihood term in Eq. \ref{eq:LUL} can also be seen as a regularizer, which makes the ranking model less overconfident when computing query likelihoods.


\section{Experiments and Discussion}
\subsection{Datasets}
We use four different publicly available answer selection datasets in our experiments: WikipassageQA \cite{cohen:2018:wikipassqa},
WikiQA \cite{yang2015:wikiqa},
InsuranceQA\_V2 \cite{feng2016:insqa},
and YahooQA \cite{tay2017:yqa}.
Statistics about the datasets are shown in Table \ref{tab:datasets}.
The four datasets also provide validation sets, 
which have size similar to the respective test sets.

\begin{table}[ht!]
\centering
\footnotesize
\scalebox{0.9}{
\begin{tabular}{lrr}
\hline
\textbf{Dataset} & \textbf{Train: \#Q (\#P/Q)} & \textbf{Test: \#Q (\#P/Q)}\\
\hline
WikiQA & 873 (9) & 243 (9) \\
WikipassageQA & 3,332 (58.3) & 416 (57.6) \\
InsuranceQA & 12,889 (500) & 2,000 (500) \\
YahooQA & 50,112 (5) & 6,283 (5) \\
\hline
\end{tabular}
}
\caption{{\small Dataset statistics. \#Q stands for \textit{number of questions} and \#P/Q is the \textit{average number of passages per question}}}\label{tab:datasets}
\end{table}

\begin{table*}[t!]
\centering
\footnotesize
\scalebox{0.9}{
\begin{tabular}{clcccccccccccc}
\hline
\textbf{ID} & \textbf{Dataset} &  \multicolumn{3}{c}{\textbf{YahooQA}} & \multicolumn{3}{c}{\textbf{WikiQA}} & \multicolumn{3}{c}{\textbf{WikipassageQA}} & \multicolumn{3}{c}{\textbf{InsuranceQA}} \\
& & MAP & MRR & P@1 & MAP & MRR & P@1  & MAP & MRR & P@1  & MAP & MRR & P@1 \\
\hline
\multicolumn{13}{c}{\textbf{Discriminative Approaches}} \\
\hline
1 & BERTSel-base \cite{li2019:bertsel}   & .942 & .942 & - & .753 & .77 & - & - & - & - & - & - & - \\
2 & BERT-PR-base \cite{xu2019:weaksup}  
& - & - & - 
& - & - & - 
& .735 & .809 &  .702 
& .413 & .496 & .401 \\
3 & BERT-PR-large  
& .965 & .965 & .939
& .844 & .856 & .765
& .775 & .838 & .748
& .410 & .492 & .394 \\
4 & BART-large  
& .967 & .967 & .943
& .845 & \textbf{.861} & .765
& .803 & \textbf{.866} & \textbf{.789}
& .435 & .518 & .423 \\
\hline
\multicolumn{13}{c}{\textbf{Generative Approaches}} \\
\hline
5 & GPT2-base [no fine-tuning]
& .499 & .499 & .265 
& .516 & .522 & .337  
& .215 & .250 & .132
& .050 & .071 & .034 \\
6 & GPT2-base$_{MLE}$ 
& .768 & .768 & .631 
& .550 & .555 & .354  
& .654 & .738 & .632
& .430 & .516 & .428 \\
\hline
7 & GPT2-base$_{LUL}$ (ours)
& .905 & .905 & .905 
& .690 & .701 & .547  
& .723 & .807 & .716
& .427 & .512 & .422 \\
8 & GPT2-base$_{RLL}$ (ours)
& .958 & .958 & .928
& .774 & .792 & .683
& .735 & .810 & .704
& .414 & .494 & .397 \\
9 & BART-base$_{LUL}$ (ours)  
& .928 & .928 & .876
& .778 & .788 & .658
& .738 & .813 & .719
& .440 & .526 & .434  \\
10 & BART-base$_{RLL}$ (ours)  
& .961 & .961 & .934
& .775 & .792 & .654
& .761 & .834 & .743
& .422 & .503 & .408  \\

\hline
11 & GPT2-large$_{LUL}$ (ours)
& .917 & .917 & .857
& .736 & .742 & .609
& .755 & .825 & .738
& .444 & .532 & .439 \\
12 & GPT2-large$_{RLL}$ (ours)
& .954 & .954 & .922
& .819 & .834 & .733
& .755 & .831 & .728
& .408 & .489 & .389 \\
13 & BART-large$_{LUL}$ (ours)
& .949 & .949 & .911
& .802 & .815 & .712
& .789 & .848 & .764
&  \textbf{.465} & \textbf{.553} & \textbf{.461}  \\
14 & BART-large$_{RLL}$ (ours)
& \textbf{.970} & \textbf{.970} & \textbf{.948}
& \textbf{.849} & \textbf{.861} & \textbf{.769}
& \textbf{.808} & \textbf{.867} & \textbf{.791}
& .444 & .529 & .433 \\
\hline
\end{tabular}
}
\caption{{\small Experimental results for different passage ranking models and datasets.}} \label{tab:results}
\end{table*}

\subsection{Language Model Setup}
We use pretrained GPT2-base (12 layers, 117M parameters), GPT2-large (24 layers, 345M params), BART-base (6 layers encoder and 6 layers decoder, 139M params) and BART-large (12 layers encoder and 12 layers decoder, 406M params) models in our experiments.
We adopted the implementation and pretrained models from \citet{wolf2019HuggingFacesTS}.
We fine-tune GPT2 and BART on each training dataset separately.
We perform a maximum of 10 fine-tuning epochs and adopt early stopping using the validation sets. Most of the hyperparmeters used for fine-tuning are the default ones from \citet{wolf2019HuggingFacesTS}\footnote{https://github.com/huggingface/transformers/blob/master /examples/run\_lm\_finetuning.py},
except for learning rate for BART, which we set to $1e-5$.

In the experiments presented below, the subscript $MLE$ corresponds to models fine-tuned using just maximum likelihood estimation (Eq. \ref{eq:MLE}),
which means that only positive examples are used.
The subscript $LUL$ corresponds to models fine-tuned using maximum likelihood and unlikelihood estimation (Eq. \ref{eq:LUL}),
while $RLL$ are models fine-tuned using the ranking loss in Eq. \ref{eq:RLL}.
For $MLE$ and $LUL$, we use a mini-batch size of 64 for InsuranceQA and 32 for the other 3 datasets.
The number of negative examples per positive examples is set to 5 in the case of $LUL$.

When fine-tuning with $RLL$ loss (Eq. \ref{eq:RLL}), we use a batch size of 8.
During training, when processing a question we randomly sample 15 negative passages from the set of negative passages of the question. However, only the negative passage with the highest score is used to update the model.
Early experiments demonstrated that this strategy performs similarly to the usual pairwise approach.

\subsection{Ranking Results}
\label{sec:results}
In Table \ref{tab:results} we present the experimental results for our proposed generative approach and four state-of-the-art discriminative baselines, 
which are based on BERT \cite{devlin2019:bert} and BART.
Both BERTSel \cite{li2019:bertsel} and BERT-PR \cite{xu2019:weaksup} fine-tuned BERT-base using a ranking loss on the score computed with $[CLS]$ token.
We trained a BERT-large model using $[CLS]$-based scoring + ranking loss (rows 3).
We additionally trained a discriminative version of BART-large (row 4) where the input for the encoder and the decoder are the passage and the question, respectively. 
As it is normally adopted in BART for classification \cite{lewis2019bart},
we take the representation generated by the decoder for the last token and use it to create a score by applying a linear layer. Such as the discriminative BERT models, we also optimize BART-large using a ranking loss.
The performance of the passage ranking models is assessed using the metrics Mean Average Precision (MAP), 
Mean Reciprocal Rank (MRR) and Precision at 1 (P@1).
Scores are computed with the official \emph{trec\_eval} tool.

In the middle part of Table \ref{tab:results},
we compare GPT2-base without any fine-tuning (row 5), and
finetuned with either $MLE$ (6), $LUL$ (7) or $RLL$ (8).
When the pretrained model only is used (no fine-tuning) the results are very poor.
Which is understandable, given that the pattern of having a passage followed by a question might not be very recurrent in the data used to pretrain GPT2.
Comparing $MLE$ (row 6) with $LUL$ (row 7),
we see that the inclusion of the unlikelihood term (Eq. \ref{eq:LUL}) has a significant positive impact for all datasets but InsuranceQA.
We believe the unlikelihood loss does not help on InsuranceQA because this dataset was not human curated and therefore contains a significant number of false-negative examples, which can hurt performance when used to compute the unlikelihood loss.
Compared to BERT-base models,
GPT2-base$_{LUL}$ is very competitive for most of the datasets except WikiQA,
while GPT2-base$_{RLL}$ demonstrates more robust results across the different datasets.
In rows 9 and 10 we show results for BART-base, where we see similar trends to GPT2-base with regard to ${LUL}$ and $RLL$ losses.
BART-base$_{RLL}$ is overall better than BART-base$_{LUL}$ and GPT2-base models.


In the bottom part of Table \ref{tab:results}, we also show results for GPT2-large and BART-large using $LUL$ and $RLL$ (rows 11 to 14).
Overall, the larger generative models do a better job than the smaller ones, as expected.
Among the generative approaches, 
BART-large$_{RLL}$ (row 14) is the model that performs the best for most of the datasets.
We believe that BART-based generative models outperform GPT2-based models due to 1) the larger number of pretraining tasks used in BART and 2) the use of bidirectional attention in the encoder side (which processes the passage).
Comparing BART-large$_{RLL}$ with discriminative BART-large (row 4),
we can see that BART-large$_{RLL}$ produces better results for InsuranceQA, while achieving similar performance for YahooQA, WikiQA and WikipassageQA.
Overall, our proposed generative approach produces state-of-the-art results on the four tested datasets in all metrics.

\begin{table}[ht!]
\centering
\footnotesize
\scalebox{0.92}{
\begin{tabular}{lcccc}
\hline
\textbf{Model} & \textbf{Likelihood} & \textbf{MAP} & \textbf{MRR} & \textbf{P@1}\\
\hline
GPT2-base$_{LUL}$   & $p_{\theta}(q|a)$ & .723 & .807 & \textbf{.716}  \\
GPT2-base$_{LUL}$   & $p_{\theta}(a|q)$ & .414 & .464 & .259  \\
GPT2-base$_{RLL}$ & $p_{\theta}(q|a)$ & \textbf{.735} & \textbf{.810} & .704  \\
GPT2-base$_{RLL}$ & $p_{\theta}(a|q)$ &  .531 & .617 & .478 \\
\hline
\end{tabular}
}
\caption{{\small Experimental results on using passage vs. question as the conditional context. Results are computed on the WikipassageQA dataset}}\label{tab:passage_likelihood}
\end{table}

\subsection{Ranking with Passage Likelihood}
A different setup that can be used for our approach is to compute the likelihood of the passage given the question, where the score for a candidate passage $a^k $ is given by $s(a_k)=p_{\theta}(a_k|q)$, and the score needs to be normalized by the passage length $|a_k|$. 
This setup is inherently more difficult because the passage is normally much longer than the question and might contain many tokens that are not relevant for the question.


\begin{table*}[t!]
\centering
\footnotesize
\scalebox{0.92}{
\begin{tabular}{ll}
\hline
\multirow{5}{*}{\textbf{Passage}} 
& This phenomenon happens usually in the winter. In 2013, Sao Paulo was the most populous city in Brazil and \\
& in  South America. According to the 2010 IBGE Census, there were 11,244,369 people residing in the city of  \\
& Sao Paulo. The census found 6,824,668 White people , 3,433,218 Pardo people , 736,083 Black people , 246,244 \\
& Asian people and 21,318 Amerindian people . In 2010, the city had 2,146,077 opposite-sex couples and 7,532  \\
& same-sex couples. The population of Sao Paulo was 52.6\% female and 47.4\% male. \\
\hline
\textbf{Generated} 
&  How diverse are the demographics of the city of Sao Paulo?  \\
\textbf{Questions} & How diverse is the population of the city of Sao Paulo?  \\
& \textit{Why are people living in the city of Sao Paulo?} \\
\hline
\hline
\multirow{5}{*}{\textbf{Passage}} 
& The presidential electors cast their ballots for President and Vice President, and Congress tallied these votes \\
& on January 8, 2009. The voter turnout for this election was broadly predicted to be high by American standards, \\
& and a record number of votes were cast. The final tally of total votes counted was 131.3 million, compared to \\
& 122.3 million in 2004 . ... . Another estimate puts the eligible voter population at 213,313,508, resulting in a \\
& turnout rate of 61.6\%, which would be the highest turnout rate since 1968. \\
\hline
& What are the demographics and voting trends of the US presidential elections? \\
\textbf{Generated} & What is a voter turnout like in the U.S. presidential elections? \\
\textbf{Questions} & \textit{What factors contributed to high voter turnout in 2008?} \\
& \textit{Why might voters have difficulty voting?} \\
\hline
\end{tabular}
}
\caption{{\small Examples of automatically generated questions using the GPT2-large$_{LUL}$ model fine-tuned on the WikipassageQA dataset with likelihood $p_{\theta}(q|a)$. The passages  were extracted from the test set.}} \label{tab:qgen}
\end{table*}


\begin{table*}[t!]
\centering
\footnotesize
\scalebox{0.895}{
\begin{tabular}{ll}
\hline
\multirow{1}{*}{\textbf{Question}} 
&  What was the election turnout for 2008?  \\
\hline
                   & The turnout in the 2008 general election was the highest in the United Kingdom since the 1920s, and the highest \\
\textbf{Generated} & turnout in the history of the UK. The turnout was higher than that for the general election of the previous year.   \\
\textbf{Passage}   & The 2008 general election was a close contest, with the Conservatives winning a majority of seats and the Labour  \\ 
                   & Party winning a majority of seats in Parliament. The Conservative Party won the largest majority in Parliament, with   \\
                   & 332 of 538 seats, and the Liberal Democrats gained the largest number of seats in the House of Commons, winning 232... \\
\hline
                   & The turnout was higher in the two states that had been held for the presidency by Barack Obama. The Democratic  \\
\textbf{Generated} & Party won the state of New Hampshire, the state that Obama won by more than 2 million votes, and the Republican \\
\textbf{Passage}   & Party won the state of Ohio by nearly 2.5 million votes. In the 2008 presidential election, Obama defeated Mitt \\ 
                   & Romney by more than 3 million votes, winning 332 of the 435 electoral votes. The election was the second of four \\
                   & presidential elections. The results of the election were announced on November 4. The election was the first in ... \\
\hline
\end{tabular}
}
\caption{{\small Examples of automatically generated passages using the GPT2-large$_{LUL}$ model fine-tuned on the WikipassageQA dataset with likelihood $p_{\theta}(a|q)$. The question was extracted from the test set.}} \label{tab:pgen}
\end{table*}

In Table \ref{tab:passage_likelihood},
we present experimental results where we compare the use of either passage or question as the conditional context.
As expected, using the likelihood of the passage given the question ($p_{\theta}(a_k|q)$) as the score results in worse performance for both fine-tuning approaches: $LUL$ and $RLL$.

\subsection{Question and Passage Generation}

A good side effect of using generative models to perform ranking is that we can use the trained model to generate new questions given a passage and vice-versa (depending on the conditioning context used for fine-tuning).
This type of synthetically generated data could be used as additional training data to improve discriminative models such as BERT-PR \cite{xu2019:weaksup}.
In Tables \ref{tab:qgen} and \ref{tab:pgen},
we present some examples of questions and passages, respectively, that were generated using our fine-tuned GPT2-large$_{LUL}$ LM.
In both cases we use a mixture of top k-sampling
\cite{fan18:topk} and nucleus sampling \cite{holtzman19:sampling} to generate the samples.
Please note that the passages in Table \ref{tab:qgen} were extracted from the test set and are not present in the training set.
The same applies for the question used in Table \ref{tab:pgen}.

In Table \ref{tab:qgen}, we can see that the generated questions are very fluent and, for most questions (except for the ones in \textit{italic}), the input passage contains the answer for the question.
In Table \ref{tab:pgen}, we can observe that the generated passages are quite related to the input question.
However, the content is normally not factual and contains inconsistencies and some repetitions.

\section{Conclusion}
We have proposed a new generative approach for IR based on large pretrained neural language models, and demonstrated their effectiveness as rankers by providing robust experimental results on four different datasets. Additionally, we demonstrated that unlikelihood-based losses are effective for allowing the use of negative examples in generative-based information retrieval.
We believe that our approach can also be effectively used for text classification problems, where the score of a class label $c$ is computed as the likelihood of generating the class label $c$ given the document $d$, $p(c|d)$.

\bibliography{anthology,emnlp2020}
\bibliographystyle{acl_natbib}

\end{document}